\documentclass{article} 
\usepackage{iclr2016_workshop,times}
\usepackage{hyperref}
\usepackage{url}
\usepackage{subfig}
\usepackage[pdftex]{graphicx}
\graphicspath{{./graphs/}}
\usepackage{natbib}
\setcitestyle{square}

\title{A Minimalistic Approach to Sum-Product Network Learning for Real Applications}

\author{Viktoriya Krakovna\\
Department of Statistics, Harvard University\\
\texttt{vkrakovna@fas.harvard.edu} \\
\And
Moshe Looks \\
Google \\
\texttt{madscience@google.com}
}

\begin{document}

\maketitle

\begin{abstract}
Sum-Product Networks (SPNs) are a class of expressive yet tractable hierarchical graphical models. LearnSPN is a structure learning algorithm for SPNs that uses hierarchical co-clustering to simultaneously identifying similar entities and similar features. The original LearnSPN algorithm assumes that all the variables are discrete and there is no missing data. We introduce a practical, simplified version of LearnSPN, MiniSPN, that runs faster and can handle missing data and heterogeneous features common in real applications. We demonstrate the performance of MiniSPN on standard benchmark datasets and on two datasets from Google's Knowledge Graph exhibiting high missingness rates and a mix of discrete and continuous features.
\end{abstract}

\section{Introduction}

\begin{figure}[t]
\centering
\begin{minipage}{.5\textwidth}
\includegraphics[scale=0.6]{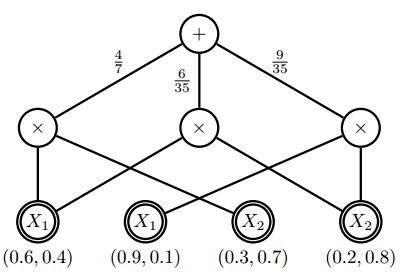}
\caption{Example of an SPN structure \\(figure from \cite{Zhao})}\label{fig:spn}
\end{minipage}%
\begin{minipage}{.5\textwidth}
\includegraphics[scale=0.17]{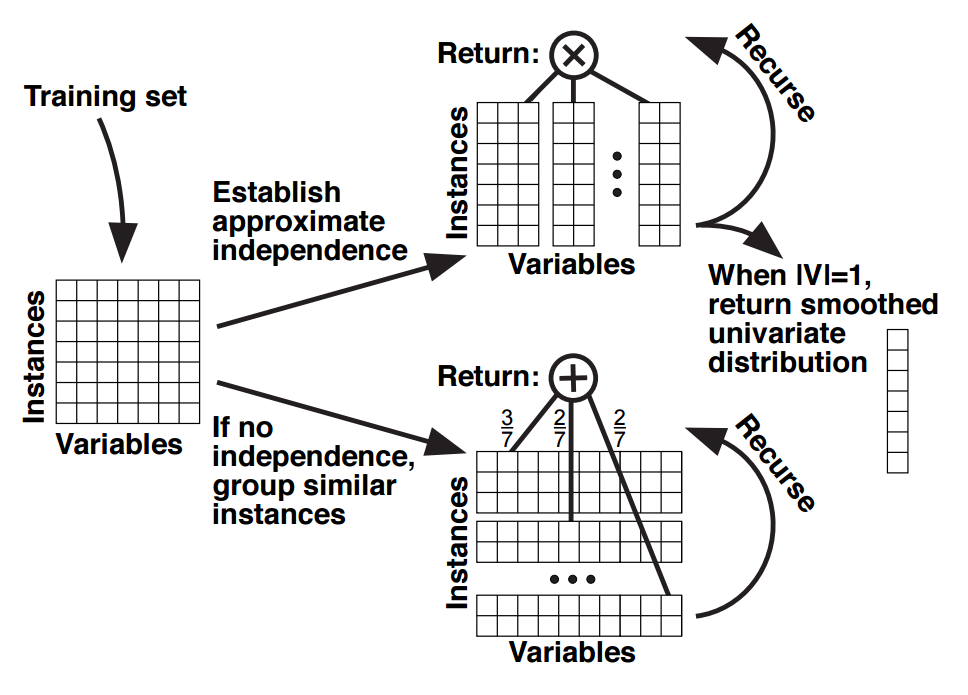}
\caption{Recursive partitioning process in the LearnSPN algorithm \\(figure from \cite{Gens})}\label{fig:learnspn}
\end{minipage}
\end{figure}

The Sum-Product Network (SPN) \citep{Poon} is a tractable and interpretable deep model. An advantage of SPNs over other graphical models such as Bayesian Networks is that they allow efficient exact inference in linear time with network size. 
An SPN represents a multivariate probability distribution with a directed acyclic graph consisting of sum nodes (clusters over instances), product nodes (partitions over features), and leaf nodes (univariate distributions over features), as shown in Figure \ref{fig:spn}.

The standard algorithms for learning SPN structure assume discrete data with no missingness, and mostly test on the same set of benchmark data sets that satisfy these criteria \citep{Rooshenas}. This is not a reasonable assumption when dealing with messy data sets in real applications. The Google Knowledge Graph (KG) is a semantic network of facts, based on Freebase \citep{Freebase}, used to generate Knowledge Panels in Google Search. KG data is quite heterogeneous, with a lot of it missing, since much more is known about some entities in the graph than others. High missingness rates can also worsen the impact of discretizing continuous variables before doing structure learning, which results in losing more of the already scarce covariance information. 

Applications like the KG are common, and there is a need for an SPN learning algorithm that can handle this kind of data. In this paper, we present MiniSPN, a simplification of a state-of-the-art SPN learning algorithm that improves its applicability, performance and speed. We demonstrate the performance of MiniSPN on KG data and on standard benchmark data sets.

\section{Variation on the LearnSPN algorithm}

LearnSPN \citep{Gens} is a greedy algorithm that performs co-clustering by recursively partitioning variables into approximately independent sets and partitioning the training data into clusters of similar instances, as shown in Figure \ref{fig:learnspn}. The variable and instance partitioning steps are applied to data slices (subsets of instances and variables) produced by previous steps. 
The variable partition step uses pairwise independence tests on the variables, and the approximately independent sets are the connected components in the resulting dependency graph.
The instance clustering step uses a naive Bayes mixture model for the clusters, where the variables in each cluster are assumed independent. The clusters are learned using hard EM with restarts, avoiding overfitting using an exponential prior on the number of clusters. 
The splitting process continues until the data slice has too few instances to test for independence, at which point all the variables in that slice are considered independent. The end result is a tree-structured SPN. 

The standard LearnSPN algorithm assumes that all the variables are discrete and there is no missing data. Hyperparameter values for the cluster penalty and the independence test critical value are determined using grid search. The clustering step seems unnecessarily complex, involving a penalty prior, EM restarts, and hyperparameter tuning. It is by far the most complicated part of the algorithm in a way that seems difficult to justify, and likely the most time-consuming due to the restarts and hyperparameter tuning. 
We propose a variation on LearnSPN called MiniSPN that handles missing data, performs lazy discretization of continuous data in variable partition step, simplifies the model in the instance clustering step, and does not require hyperparameter search. 

\begin{table}[t]
\centering
\small
\caption{Average log likelihood and runtime comparison on KG data sets (best performing methods are shown in bold).}
\label{kg-results}
\begin{tabular}{l|lll|lll}\hline
& \multicolumn{3}{c}{Test set log likelihood} & \multicolumn{3}{|c}{Runtime (seconds)} \\\hline
Data set         & Pareto  & Hybrid          & MiniSPN         & Pareto   & Hybrid   & MiniSPN         \\\hline
Professions-10K  & -10.2   & -6.2            & \textbf{-6.09}  & 5.3      & 3.7      & \textbf{0.4}    \\
Professions-100K & -6.61   & -6.53           & \textbf{-6.44}  & 72       & 131      & \textbf{7.2}    \\
Dates-10K        & -8.66   & \textbf{-8.53}  & -8.68           & 1.7      & 2.4      & \textbf{0.26}   \\
Dates-100K       & -17.1   & -16.7           & \textbf{-16.5}  & 29       & 566      & \textbf{5.4}  \\\hline 
\end{tabular}
\vspace{1em}
\centering
\caption{Average log likelihood and runtime comparison on literature data sets (best performing methods are shown in bold).}
\label{lit-results}
\small
\begin{tabular}{l|llll|llll}\hline
& \multicolumn{4}{c}{Test set log likelihood} & \multicolumn{4}{|c}{Runtime (seconds)} \\\hline
Data set               & Pareto & Hybrid         & MiniSPN        & LearnSPN        & Pareto       & Hybrid & MiniSPN      & LearnSPN \\\hline
NLTCS        & -6.33  & \textbf{-6.03} & -6.12          & -6.1            & \textbf{4.8} & 35     & \textbf{1.4} & 60       \\
MSNBC        & -6.54  & -6.4           & -6.61          & \textbf{-6.11}  & 61           & 212    & \textbf{5.6} & 2400     \\
KDDCup  & -2.17  & \textbf{-2.13} & \textbf{-2.14} & -2.21           & 152          & 2080   & \textbf{23}  & 400      \\
Plants       & -17.3  & \textbf{-13.1} & \textbf{-13.2} & \textbf{-13}    & 28           & 780    & \textbf{11}  & 160      \\
Audio        & -41.9  & \textbf{-39.9} & \textbf{-40}   & -40.5           & 28           & 556    & \textbf{12}  & 955      \\
Jester       & -54.6  & \textbf{-52.9} & \textbf{-53}   & -53.4           & 13           & 193    & \textbf{6.7} & 1190     \\
Netflix      & -59.5  & \textbf{-56.7} & \textbf{-56.8} & -57.3           & 27           & 766    & \textbf{14}  & 1230     \\
Accidents    & -40.4  & -32.5          & -32.6          & \textbf{-30.3}  & 31           & 1140   & \textbf{18}  & 330      \\
Retail       & -11.1  & \textbf{-11}   & -11.1          & \textbf{-11.09} & 25           & 63     & \textbf{7.3} & 100      \\
Pumsb-star   & -40.8  & -28.4          & -28.3          & \textbf{-25}    & 47           & 1100   & \textbf{22}  & 350      \\
DNA          & -98.1  & -91.5          & -93.9          & \textbf{-89}    & \textbf{6.3} & 45     & \textbf{3}   & 300      \\
Kosarek      & -11.2  & \textbf{-10.8} & \textbf{-10.9} & \textbf{-11}    & 90           & 537    & \textbf{22}  & 200      \\
MSWeb        & -10.7  & \textbf{-9.94} & \textbf{-10.1} & -10.26          & 75           & 572    & \textbf{34}  & 260      \\
Book         & -35.1  & \textbf{-34.7} & \textbf{-34.7} & -36.4           & 83           & 181    & \textbf{32}  & 350      \\
EachMovie    & -55    & \textbf{-52.3} & \textbf{-52.2} & -52.5           & 62           & 218    & \textbf{22}  & 220      \\
WebKB        & -161   & \textbf{-155}  & \textbf{-155}  & -162            & 37           & 169    & \textbf{38}  & 900      \\
Reuters-52   & -92    & \textbf{-85.2} & \textbf{-84.7} & -86.5           & \textbf{76}  & 656    & 95           & 2900     \\
Newsgroup & -156   & \textbf{-152}  & \textbf{-152}  & -160.5          & 181          & 1190   & \textbf{139} & 28000    \\
BBC          & -258   & \textbf{-250}  & \textbf{-249}  & \textbf{-250}   & \textbf{33}  & 123    & \textbf{42}  & 900      \\
Ad           & -52.3  & -49.5          & -49.2          & \textbf{-22}    & \textbf{58}  & 92     & \textbf{50}  & 300        \\\hline
\end{tabular}
\end{table}

We simplify the naive Bayes mixture model in the instance clustering step by attempting a split into two clusters at any given point, and eliminating the cluster penalty prior, which results in a more greedy approach than in LearnSPN that does not require restarts or hyperparameter tuning. This seems like a natural choice of simplification - an extension of the greedy approach used at the top level of the LearnSPN algorithm.
We compare a partition into univariate leaves to a mixture of two partitions into univariate leaves (generated using hard EM), and the split succeeds if the two-cluster version has higher validation set likelihood. 
If the split succeeds, we apply it to each of the two resulting data slices, and only move on to a variable partition step after the clustering step fails. The greedy approach is similar to the one used in the SPN-B method \citep{Vergari}, which however alternates between variable and instance splits by default, thus building even deeper SPNs. 

In the variable partition step, we perform an independence test using the subset of rows where both variables are not missing, and conclude independence if the number of such rows is below threshold. 
We apply binary binning to each continuous variable, using its median in the data slice as a cutoff. 


We compare to the ``Pareto" algorithm, previously used for learning SPN models in KG, inspired by the work of \cite{Grosse}. 
It produces a Pareto-optimal set of models, trading off between degrees of freedom and validation set log likelihood score. At each iteration, production rules are randomly applied to add partition and mixture splits to the models in the current model set, and the new models are added to the model set. If a model in the model set has both lower degrees of freedom and higher log likelihood score than another model, the inferior model is removed from the set. The algorithm returns the model from the Pareto model set with the highest validation log likelihood. We also compare to a hybrid method, with the Pareto algorithm initialized by MiniSPN.

\section{Summary of experiments}
We use two data sets from the Knowledge Graph People collection. In the KG Professions data set, most of the variables are boolean indicators of whether each person belongs to a particular profession. There are 83 boolean variables and 4 continuous variables. In the KG Dates data set, there are 14 continuous variables representing dates of life events for each person and their spouse(s), with around 95\% of the data missing. We use subsets of 10000 and 100000 instances from each of these data sets, and randomly split the data sets into a training and test set.

On the KG data sets, we compare MiniSPN, Pareto and Hybrid algorithms. We were not able to apply the standard LearnSPN algorithm on these data sets, since they contain missing data. Table \ref{kg-results} shows log likelihood performance on the test set and runtime performance. MiniSPN does better than Pareto, both in terms of log likelihood and runtime. Hybrid performs comparably to MiniSPN, but is usually the slowest of the three. 

We use 20 benchmark data sets from the literature (exactly the same ones used in the
LearnSPN paper \citep{Gens}) to compare the performance of MiniSPN with the standard LearnSPN algorithm. We are particularly interested in the effect of MiniSPN's simple two-cluster instance split relative to the more complex instance split with the exponential prior and EM restarts used in the standard LearnSPN. Table \ref{lit-results} shows log likelihood performance on the test set and runtime performance. Like on the KG data, we find that MiniSPN uniformly outperforms Pareto, and performs similarly to Hybrid and LearnSPN but runs much faster (on the most time-intensive data set, Newsgroup, MiniSPN takes 2 minutes while LearnSPN takes 8 hours). 

\section{Conclusion}

Sum-product networks have been receiving increasing attention from researchers due to their expressiveness, efficient inference and interpretability, and many learning algorithms have been developed in the past few years. While recent developments have mostly focused on improving performance on benchmark data sets, our variation on a classical learning algorithm is simple yet has a large impact on usability, by improving speed and making it possible to apply to messy real data sets. 

\bibliography{spn_bibliography}

\end{document}